\documentclass[conference]{IEEEtran}
\IEEEoverridecommandlockouts
\usepackage{cite}
\usepackage{amsmath,amssymb,amsfonts}
\usepackage{algorithmic}
\usepackage{graphicx}
\usepackage[numbers, square]{natbib}
\usepackage{textcomp}
\usepackage{booktabs}
\usepackage{graphicx}
\usepackage{array}
\usepackage{url}
\usepackage{booktabs}

\usepackage{todonotes}
\begin{document}

\title{GeomCLIP: Contrastive Geometry-Text \\ Pre-training for Molecules
}

\author{\IEEEauthorblockN{Teng Xiao$^\dagger$, Chao Cui$^{\ddagger}$, Huaisheng Zhu$^\dagger$,  Vasant G. Honavar$^\dagger$}
\IEEEauthorblockA{$^\dagger$\textit{Artificial Intelligence Research Laboratory}, The Pennsylvania State University\\
$^\ddagger$\textit{Tsinghua Shenzhen International Graduate School}, Tsinghua University\\
\{tengxiao, hvz5312, vhonavar\}@psu.edu chaocui01@gmail.com }
}

\maketitle

\begin{abstract}
Pretraining molecular representations is crucial for drug and material discovery. Recent methods focus on learning representations from geometric structures, effectively capturing 3D position information. Yet, they overlook the rich information in biomedical texts, which detail molecules' properties and substructures. With this in mind, we set up a data collection effort for 200K pairs of ground-state geometric structures and biomedical texts, resulting in a \texttt{PubChem3D} dataset. Based
on this dataset, we propose the \texttt{GeomCLIP} framework to enhance for multi-modal representation learning from molecular structures and biomedical text. During pre-training, we design two types of tasks, i.e., multimodal representation alignment and unimodal denoising pretraining, to align the 3D geometric encoder with textual information and, at the same time, preserve its original representation power. Experimental results show the effectiveness of \texttt{GeomCLIP} in various tasks such as molecular property prediction, zero-shot text-molecule retrieval, and 3D molecule captioning. 
Our code and collected dataset are available at \url{https://github.com/xiaocui3737/GeomCLIP}.
\end{abstract}


\section{INTRODUCTION}
\begin{figure*}[t!] 
\centering 
\includegraphics[width=0.835\textwidth]{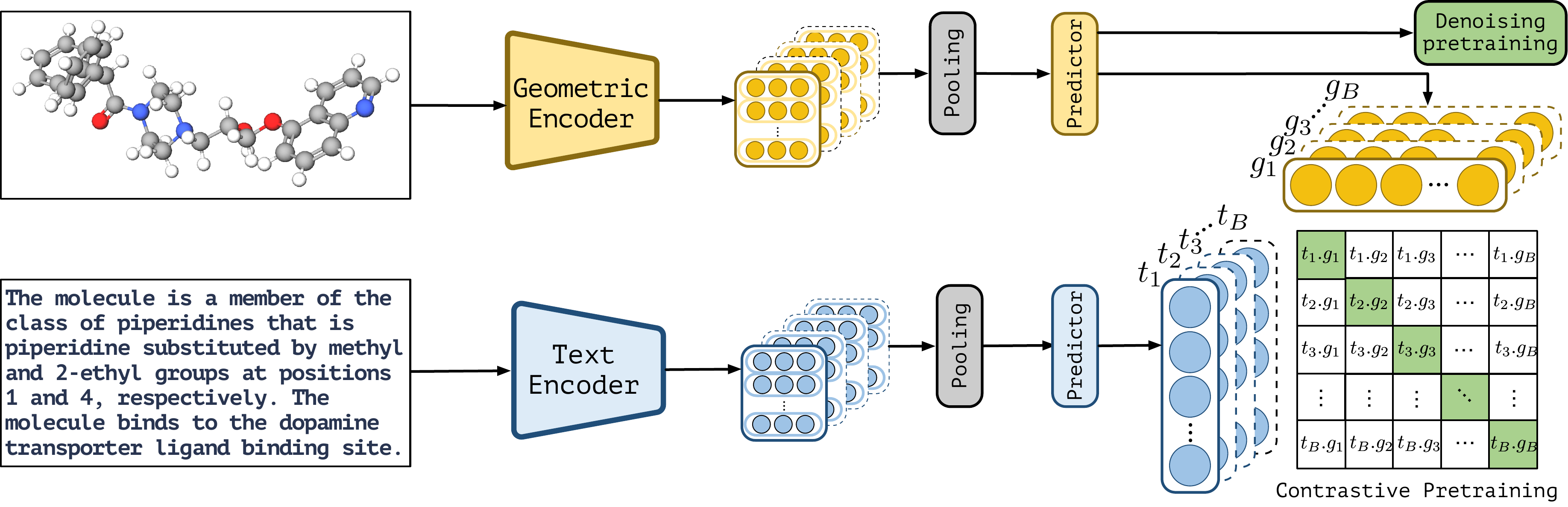}
\vskip -1.1em
\caption{\small Framework of \texttt{GeomCLIP}: A dual-encoder pre-training scheme for 3D geometric molecules and their text observations.}
\label{fig:framework} 
\vskip -1.5em
\end{figure*}
The problem of learning useful image~\cite{radford2021learning,singh2022flava}, video~\cite{luo2022clip4clip}, and audio~\cite{guzhov2022audioclip} representations by incorporating text supervision~\cite{radford2021learning} has been extensively studied in the literature. Because of the practical importance of generating molecular structures from textual descriptions of molecular characteristics, substructures, etc., there is a growing interest in multimodal representation learning from molecular structures and biomedical text ~\cite{ edwards2022translation,liu2023multi,liu2023molca,fang2023mol,seidl2023enhancing,xiao2024Cal}. 

Existing works for multimodal learning on molecules and texts operate in two ways: (1) Sequence-based methods model molecules as 1D sequences, such as SMILES~\cite{pei2023biot5,liu-etal-2023-molxpt,christofidellis2023unifying,edwards2022translation}; (2) Graph-based methods aim to capture the 2D structures in molecules~\cite{su2022molecular,edwards2021text2mol,liu2023molca,xiao2024simple,chen2022ba,xiao2024molbind,zhu20243m,xiaoefficient}. These methods, however, do not investigate the effect of 3D geometric structures, which largely determine the physical and chemical properties of molecules~\cite{liu2023group,liu2022molecular}. The learning of geometric representation
for molecules is critical in various applications for quantum chemistry~\cite{gilmer2017neural}, protein structure prediction~\cite{schutt2021equivariant}, materials science~\cite{schmidt2019recent}, and drug discovery~\cite{stokes2020deep}. 
Thus, a promising direction is to pretrain molecular representations based on 3D geometric structures and text descriptions, which is the main focus of this paper. 
Concurrent to our work, Tang \textit{et al.} and Li \textit{et al.}~\cite{tang2023mollm,li2024towards} leverage 3D information to aid in multimodal learning. However, they rely directly on RDKit~\cite{landrum2013rdkit} to generate approximate 3D geometries from SMILES for evaluation and training, which are not ground-state geometries, are ambiguous, and can introduce significant noise~\cite{xu2021geodiff,xu2021molecule3d}. In addition, their heavy emphasis on cross-modal alignment causes them to overlook the unimodal information of 3D geometric structures, as demonstrated in our experiments.

In this work, we pioneer the creation of a high-quality dataset called \texttt{PubChem3D}, comprising \texttt{203,257} pairs of ground-state geometric structures and biomedical texts. While multimodal datasets exist in other domains~\cite{radford2021learning,guzhov2022audioclip}, a conspicuous gap remains when it comes to 3D molecule-text. This gap stems from two main challenges: (i) acquiring ground-state geometric structures is costly due to the time-intensive nature of methods such as density functional theory (DFT)~\cite{parr1979local}. (ii) Annotating the text of molecules is expensive given the need for depth of expert knowledge.
To address these challenges, we describe a data collection effort, manually collecting molecule-text pairs from various licensed sources (see \S~\ref{sec:datasets} for details). To further enhance the representation learning of molecules through text, 
we propose a simple \texttt{GeomCLIP} framework, inspired by CLIP~\cite{radford2021learning} to perform multimodal pretraining of 3D geometry structures and textual descriptions of molecules based on collected \texttt{PubChem3D}.
As shown in Figure~\ref{fig:framework},  \texttt{GeomCLIP} comprises two encoders, each tailored to learn 3D geometry or text representations of molecules. The geometry and text encoders are aligned via a task-agnostic joint contrastive objective to predict correct pairings within a batch of (geometry, text) pairs. \texttt{GeomCLIP} also employs a denoising objective to maintain the original effectiveness of the geometric encoder in capturing the 3D positions of molecules. Extensive experiments show the effectiveness of \texttt{GeomCLIP} in various downstream tasks such as  molecular property prediction, text-molecule retrieval, and molecule captioning. 

Our key contributions include: (1) We study a novel problem of learning multi-modal molecular representations, integrating 3D geometry with textual data. (2) We meticulously curate and develop a dataset combining text with ground-state 3D geometry of molecules. Building upon our dataset, we propose a straightforward yet highly effective approach, \texttt{GeomCLIP} for enhancing 3D geometry representations of molecules. (3) Extensive experiments show that our \texttt{GeomCLIP} yields improved  performance  on various downstream tasks such as 3D molecular property prediction, text-to-molecule  (and back) retrieval, and 3D molecule captioning. 

\section{Related Work}
\subsection{Molecule Representation Learning}
Molecular representation learning plays a crucial role in fields such as drug discovery~\cite{shen2019molecular} and material design~\cite{pollice2021data}. In particular, self-supervised pretraining approaches applied to molecular data have exhibited promising performance without the need for labels. Some works focus on 1D SMILES strings~\cite{krenn2020self} and 2D molecular graphs~\cite{wang2022improving}, leveraging sequence-based or graph-based pretraining methods to learn molecular representations effectively. However, they overlook the 3D geometric structure of molecules, which is vital since the physical and chemical properties of the molecule are significantly influenced by its 3D geometry~\cite{koukos2019protein,schutt2018schnet}. Thus, recent research focuses on learning representations from 3D geometric graphs through self-supervised methods~\cite{liu2021pre,feng2023fractional,xiao2021general}. Different from these studies, which neglect the semantic text information of molecules. Our work aims to enhance geometry representation of molecules by using textual descriptions.

\subsection{Text-Molecule Multi-modal Learning}
It has been broadly studied how to learn better image~\cite{radford2021learning}, video~\cite{luo2022clip4clip}, and audio~\cite{guzhov2022audioclip} representations by incorporating text supervision~\cite{radford2021learning}. Since natural language enables nuanced expression of molecular characteristics, substructures, and biomedical understanding, multi-modal representation learning on molecule and biomedical text has recently attracted considerable attention~\cite{su2022molecular, edwards2021text2mol,liu2023multi,liu2024git,liu2023molca}. 
Existing works for multi-modal learning on molecules in two ways: (1) Sequence-based
methods model molecules on 1D sequences~\cite{pei2023biot5,liu-etal-2023-molxpt,edwards2022translation}; (2) Graph-based methods seek to capture 2D structures in molecules~\cite{su2022molecular,edwards2021text2mol,liu2023molca,luo2023molfm,xiao2021learning,xiao2022decoupled}. However,
these two lines of work investigate the effect of 3D geometric structures less.  Concurrent to our work, ~\cite{tang2023mollm,li2024towards} leverage 3D information to help multi-modal learning. However, the 3D positions serve only as auxiliary information and cannot be used for supervision during alignment. Furthermore, they rely directly on RDKit~\cite{landrum2013rdkit} to generate 3D geometries from SMILES, which can be inaccurate and introduce significant noise~\cite{xu2021molecule3d,xu2021geodiff}. Consequently, their performance is suboptimal, as demonstrated in our experiments. In contrast, our work pioneers the creation of a dataset PubChem3D, comprising pairs of ground-state geometric structures and biomedical texts, and enhances the geometric representation learning of molecules through text.







\section{METHODS}

\subsection{Dataset Construction--\texttt{\texttt{PubChem3D}}}
\label{sec:datasets}
\noindent \textbf{3D Geometry collection.} To build our dataset, we consider two large databases containing ground-state geometries obtained by DFT computations. The first is PubChemQC~\cite{nakata2017pubchemqc}, which contains over 3.9 million molecules, including their molecular graphs and ground-state 3D geometries. Furthermore, we also consider GEOM~\cite{axelrod2022geom}, a database of high-quality geometries for 430,000 molecules. 
Each molecule in GEOM has multiple
geometric structures. As the top-10 conformers are sufficient to cover most
conformers with an equilibrium state, we sample the top-10 geometry structures for each molecule with the highest possibility and lowest energy. We merge these two databases and extract the IUPAC International Chemical Identifier (InChI)~\cite{heller2015inchi} for each molecule.

\noindent \textbf{Text annotation collection.}
Based on the 3D geometries collected, we aim to gather their text annotations. Manual annotation of molecules is costly due to its complexity. Hence, we turn to PubChem~\cite{kim2021pubchem}, a freely accessible and essential resource for chemical research, for comprehensive and authoritative molecular text annotations. PubChem contains text descriptions for many molecules, submitted by various research institutions. Using the InChI as a unique identifier, we retrieve text descriptions from PubChem for each geometry structure collected. The text annotations include descriptions related to properties or 3D geometry information. For instance, in molecule with CID 444795: “The molecule is a retinoic acid in which all four exocyclic double bonds have E- (trans-) geometry", and (2) In molecule with CID 5375200:" The molecule is an abscisic acid in which the two acyclic double bonds both have trans-geometry”.

\noindent \textbf{Final dataset.} The final dataset \texttt{PubChem3D} contains \texttt{203,257} geometry-text pairs, in which \texttt{70,981}  and \texttt{132,276} ground-state geometries come from PubChemQC and GEOM, respectively.
We counted the types of molecular descriptions in the collected dataset, such as toxicity, solubility, and color in Table~\ref{table:volume}, which shows the diversity of texts.

\begin{table}
\caption{Data volume, average number of atoms, and word counts of source datasets}
\vskip -1.1em
\label{table:volume}
\begin{tabular}{lccc}
\toprule
\textbf{Data Source} & \textbf{Quantity} & \textbf{Average Heavy Atoms} &  \textbf{Average Words} \\
\midrule
PubChemQC & 70981 & 13.79 & 68.54 \\
\midrule
GEOM-Drug & 132276 & 25.58 & 31.81 \\
\midrule
GEOM-QM9 & 133885 & 8.80 & —\\
\midrule
PubChem3D & 203257 & 21.46 & 44.64 \\
\bottomrule
\end{tabular}
\vskip -1.1em
\end{table}

\subsection{Joint Modeling of  Geometries and Texts}
\vspace{-0.2em}
\noindent \textbf{Modality Alignment Task.} Our work introduces a simple geometry-text pretraining approach, \texttt{ GeomCLIP}, which enables advanced cross-modal representation. The alignment objective is inspired by the fact that both the semantic text representation and the 3D geometric representation of the same molecule should be as close to one another as possible. 
We align the embedding of corresponding geometry-text pairs while distancing other pairs in the same batch:
{
\setlength{\abovedisplayskip}{5pt}
\setlength{\belowdisplayskip}{5pt}
\begingroup\makeatletter\def\f@size{8.5}\check@mathfonts\def\maketag@@@#1{\hbox{\m@th\normalfont\normalfont#1}}
\begin{equation}
\label{eq:contrast1}
\mathcal{L}_{\text{con}}=-\frac{1}{|\mathcal{B}|}\sum_{(g_{i}, t_{i}) \in \mathcal{B}}\log \operatorname{NCE}(\boldsymbol{g}_i, \boldsymbol{t}_i)+\log \operatorname{NCE}(\boldsymbol{t}_i, \boldsymbol{g}_i),
\end{equation}
\endgroup}
Where $\mathcal{B}$ represents the batch of geometry-text pairs, $\operatorname{NCE}(\boldsymbol{g}_i, \boldsymbol{t}_i)$ and $\operatorname{NCE}(\boldsymbol{t}_i, \boldsymbol{g}_i)$ denote the contrastive losses for geometry-to-text and text-to-geometry similarities, respectively. $\boldsymbol{g}_i$ and $\boldsymbol{t}_i$ are the representations from the geometric encoder $f_{\theta}$ and the text encoder $f_{\phi}$, detailed in Appendix~\ref{sec:model}, respectively. The geometry-to-text contrastive loss, $\operatorname{NCE}(\boldsymbol{g}_i, \boldsymbol{t}_i)$, describes the likelihood of correctly ranking the molecules given its text.
{
\setlength{\abovedisplayskip}{5pt}
\setlength{\belowdisplayskip}{5pt}
\begingroup\makeatletter\def\f@size{8.7}\check@mathfonts\def\maketag@@@#1{\hbox{\m@th\normalfont\normalfont#1}}
\begin{equation}
\label{eq:contrast}
    \operatorname{NCE}(\boldsymbol{g}_i, \boldsymbol{t}_i)=\log \frac{\exp ( \cos (\boldsymbol{g}_i, \boldsymbol{t}_i) / \tau)}{\sum_{j=1}^{|\mathcal{B}|} \exp ( \cos (\boldsymbol{g}_{i}, \boldsymbol{t}_{j}) / \tau)}, 
\end{equation}
\endgroup}
\noindent where $\tau$ is temperature and $\boldsymbol{t}_{i}$ represents positive text embeddings that overlap with  3D geometry embedding.  $\boldsymbol{t}_{j}$ is negative text embedding implicitly formed by other text embeddings in the batch. A symmetric equivalent, text-geometry contrastive loss $\operatorname{NCE}(\boldsymbol{t}_i, \boldsymbol{g}_i)$ can be similarly calculated.\\
\noindent \textbf{Denoising Pretaining Task}. 
To  preserve the unimodal information of 3D molecules when injecting the cross-modality information from biomedical texts,  we incorporate a denoising pretraining task on geometric encoder that enables the model to effectively capture 3D structural information during alignment. Specifically, given an input 3D molecule $G$, we perturb it by adding i.i.d. Gaussian noise to its atomic positions $\boldsymbol{p}_i$ and masking atoms, resulting in a noisy version of the molecule:
{
\setlength{\abovedisplayskip}{5pt}
\setlength{\belowdisplayskip}{5pt}
\begingroup\makeatletter\def\f@size{8.7}\check@mathfonts\def\maketag@@@#1{\hbox{\m@th\normalfont\normalfont#1}}
\begin{align}
&\tilde{G}=\left\{(\tilde{\boldsymbol{p}}_{1},{\boldsymbol{x}}_1) \ldots,( \tilde{\boldsymbol{p}}_{N},{\boldsymbol{x}}_N)\right\}, \\
&\text { where } \tilde{\boldsymbol{p}}_i=\boldsymbol{p}_i+\sigma \boldsymbol{\epsilon}_i \text { and } \boldsymbol{\epsilon}_i \sim \mathcal{N}(0, I), \nonumber 
\end{align}
\endgroup}
where noise $\sigma=1$. We also randomly mask 15\% of the atoms in the molecule~\cite{zhou2022uni}. We denote the raw masked molecule and the noised complementary molecule by ${G}[m]$ and $\tilde{G}[1-m]$, respectively, where $m$ is the binary index of the masked atoms. The task is formulated as a denoising autoencoder to predict  noised coordinates and types of masked atoms:
{
\setlength{\abovedisplayskip}{5pt}
\setlength{\belowdisplayskip}{5pt}
\begingroup\makeatletter\def\f@size{8.7}\check@mathfonts\def\maketag@@@#1{\hbox{\m@th\normalfont\normalfont#1}}
\begin{equation}
\mathcal{L}_{\text{denoising}}=\sum_{i=1}^{|\mathcal{D}|}\|f_{\psi}(\tilde{\boldsymbol{g}}_{i})-{G}_{i}[m]\|^2,
\end{equation}
\endgroup
}
$\tilde{\boldsymbol{g}}_{i}=f_{\theta}(\tilde{G}_{i}[1-m])$ is the representation of noised complementary molecule, and $f_{\psi}$ is the decoder proposed  in~\cite{zhou2022uni}.\\
\noindent \textbf{Overall  Objective:} 
To promote representation alignment and maintain the capacity to capture geometric information, \texttt{GeomCLIP} simultaneously minimizes two losses:
{
\setlength{\abovedisplayskip}{5pt}
\setlength{\belowdisplayskip}{5pt}
\begingroup\makeatletter\def\f@size{10}\check@mathfonts\def\maketag@@@#1{\hbox{\m@th\normalfont\normalfont#1}}
\begin{align}
\mathcal{L}_{\text{GeomCLIP}}=\mathcal{L}_{\text{con}}+ \alpha \mathcal{L}_{\text{mask}},
\end{align}
\endgroup
}
where $\alpha$ is the loss weighting hyperparameter.

\begin{figure}[t!] 
\centering 
\includegraphics[width=0.495\textwidth]{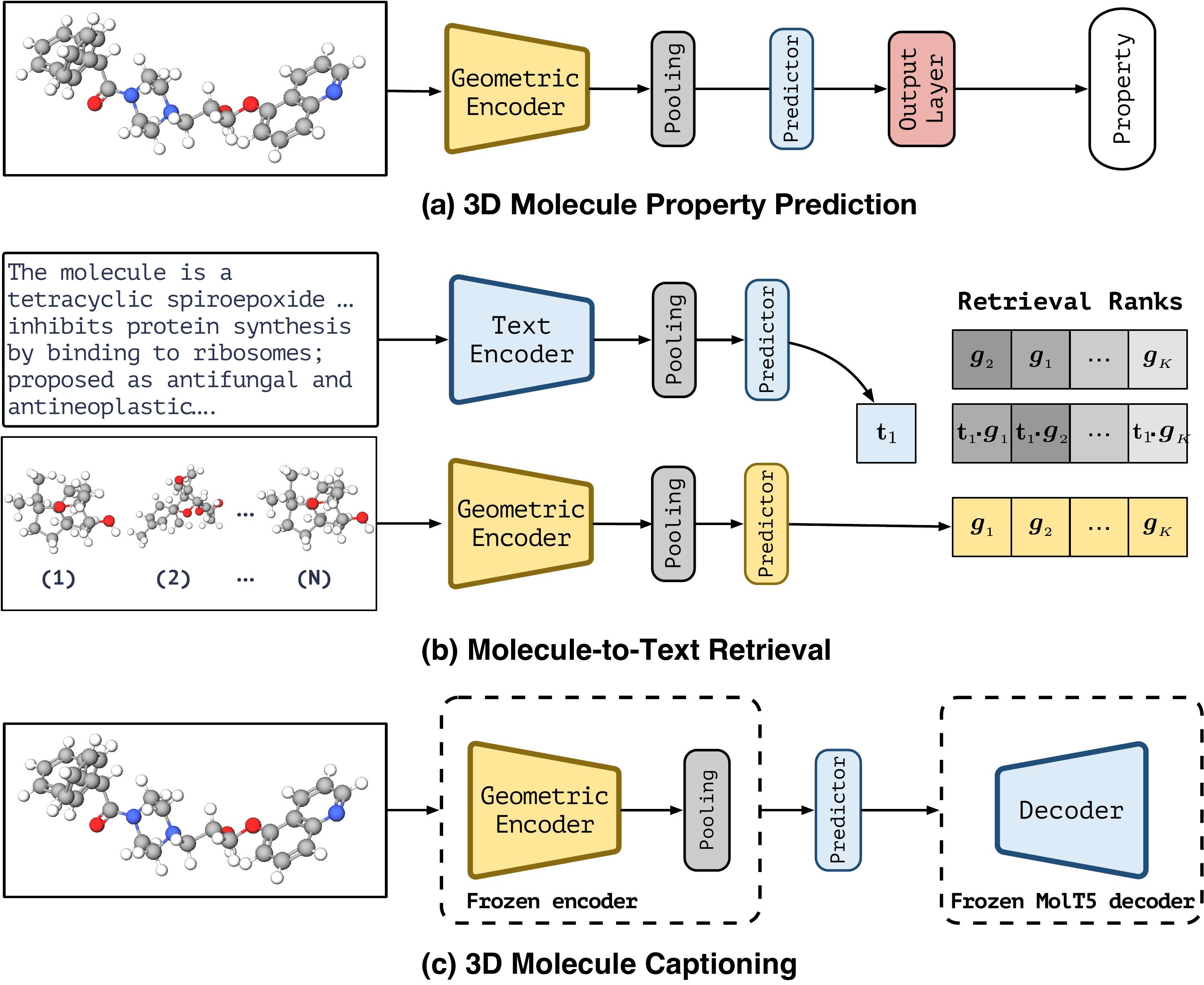}
\caption{$\texttt{GeomCLIP}$ can perform different downstream tasks: (a) molecular property prediction, where $\texttt{GeomCLIP}$ is fine-tuned to predict properties of molecules. (b) Pretrained geometric and text encoders can perform zero-shot molecule-text retrieval without any annotations. (c) Molecule captioning, where we integrate $\texttt{GeomCLIP}$’s aligned molecule representation with the MolT5 pretrained text decoder through optimizing predictor.}
\label{fig:task} 
\end{figure}

\begin{table*}[t]
\centering
\fontsize{9}{9}\selectfont
\caption{
\small
Results on 12 quantum mechanics prediction tasks from QM9 and the best results are marked in \textbf{bold}.
}
\vspace{-2ex}
\label{tab:main_QM9_result}
\resizebox{0.95\textwidth}{!}{
\begin{tabular}{l c c c c c c c c c c c c}
\toprule
Model & $\alpha$ $\downarrow$ & $\nabla \mathcal{E}$ $\downarrow$ & $\mathcal{E}_\text{HOMO}$ $\downarrow$ & $\mathcal{E}_\text{LUMO}$ $\downarrow$ & $\mu$ $\downarrow$ & $C_v$ $\downarrow$ & $G$ $\downarrow$ & $H$ $\downarrow$ & $R^2$ $\downarrow$ & $U$ $\downarrow$ & $U_0$ $\downarrow$ & ZPVE $\downarrow$\\
\midrule

3D InfoMax & 0.057 & 42.09 & 25.90 & 21.60 & 0.028 & 0.030 & 13.73 & 13.62 & 0.141 & 13.81 & 13.30 & 1.670\\

GraphMVP & 0.056 & 41.99 & 25.75 & {21.58} & {0.027} & {0.029} & 13.43 & 13.31 & 0.136 & 13.03 & 13.07 & 1.609\\

MoleculeSDE &0.054& 41.77& 25.74 &21.41 &0.026 &0.028 &13.07 &12.05 &0.151 &12.54 &12.04 &1.587 \\

MoleculeJAE &0.056& 42.73& 25.95& 21.55& 0.027 &0.029 &11.22& 10.70 &0.141& 10.81& 10.70& 1.559\\

3D-MoLM &0.055& 42.53& 24.76 & 21.39& 0.023 &0.026 &12.51& 11.55 &0.135& 10.78& 11.22& 1.468\\

Uni-Mol & 0.051 & 41.01& 23.31 & 20.75 & \textbf{0.016} & 0.023 & 9.52 & 8.73 &0.128 & 9.77 & 9.65& 1.345\\

\midrule

\texttt{GeomCLIP} & \textbf{0.048} & \textbf{39.52} &  \textbf{22.78} & \textbf{19.61} & \textbf{0.016} & \textbf{0.022} & \textbf{8.61} & \textbf{7.49} & \textbf{0.118} &  \textbf{8.27} &  \textbf{8.19} & \textbf{1.209}\\
\bottomrule
\end{tabular}}
\vspace{-1.5em}
\end{table*}

\subsection{Model Architecture}
\label{sec:model}
\noindent \textbf{3D Molecular Encoder.} 
The 3D molecular encoder $f_{\theta}$ in GeomCLIP draws inspiration from recent advancements in
transformer-based models for encoding geometry information, as demonstrated in works~\cite{zhou2022uni,luo2022one}. In this work, 
we employ Uni-Mol\footnote{\url{https://github.com/dptech-corp/Uni-Mol/tree/main/unimol}}~\cite{zhou2022uni} as our 3D molecular encoder,
which is a transformer-based model with two inputs, atom types and atom 3D coordinates. The atom representation is initialized from atom types, by the embedding layer, and the representation for each pair of atoms is initialized using invariant spatial positional encoding from 3D coordinates. 
Then, the representations of atoms and atom pairs communicate with each other in the self-attention module. Formally, Uni-Mol $f_{\text{geom}}$ performs 3D encoding steps to obtain sequential atomic representations:
{
\begingroup\makeatletter\def\f@size{9.5}\check@mathfonts\def\maketag@@@#1{\hbox{\m@th\normalfont\normalfont#1}}
\begin{align}
\left[\boldsymbol{z}_1, \boldsymbol{z}_2, \ldots, \boldsymbol{z}_{N}\right]=f_{\text{geom}}(G), \\
~~~\overline{\boldsymbol{z}}=\text{pooling}(\left[\boldsymbol{z}_1, \boldsymbol{z}_2, \ldots, \boldsymbol{z}_{N}\right]),
\end{align}
\endgroup}
where $\boldsymbol{z}_i$ corresponds to the representation of the $i$-th atom and we conduct a mean pooling operation to get the global representation of the molecule. We then use a projection MLP layer to map from the encoder’s representation to the multimodal embedding space: $\boldsymbol{g}=f_{\text{proj}}(\overline{\boldsymbol{z}})$. The predictor customizes molecular representation for distinct pretraining tasks while sharing the previous backbone encoder, and acts as a natural task-specific adapter. The whole process can also be expressed as $\boldsymbol{g} = f_{\theta}(G)$ signifying the outcome
of the encoding operation within the input 3D geometric graph.
\noindent \textbf{Biomedical Text Encoder.}  The text encoder’s foundation is rooted in the recent advances of transformer-based models to encode scientific textual descriptions into latent spaces.
To inject potentially useful scientific knowledge from the literature into the text encoder, we initialize it with the Sci-BERT’s checkpoint\footnote{\url{https://huggingface.co/allenai/scibert_scivocab_uncased}}~\cite{beltagy2019scibert} at denoted as $f_{\text{text}}(T)$, which is a transformer-based encoder pretrained on scientific publications. We utilize the pooling representation of the \texttt{[CLS]} token in Sci-BERT as whole text representations: $\boldsymbol{t}_{\text{CLS}}=f_{\text{text}}(T)$[CLS]. Similar to molecule, we also introduce an additional MLP predictor to map the text representation to multimodal space: The encoding process can also be expressed as $\boldsymbol{t}=f_{\phi}(T)$.





\begin{table}[t]
    \centering
    \setlength{\tabcolsep}{9pt}
    \fontsize{9pt}{9}\selectfont
    \caption{\small Molecule-Text retrieval performances (\%). $\dag$ denotes method which is also pretained  on our \texttt{PubChem3D}.}
        \label{tab:retrieve_PubChem3D}
    \resizebox{0.47\textwidth}{!}{
    \begin{tabular}{lcccc} \toprule
                   & \multicolumn{2}{c}{Molecule2Text}                 & \multicolumn{2}{c}{Text2Molecule}                 \\ \cmidrule(lr){2-3} \cmidrule(lr){4-5}
Model         & Acc  & R@20 & Acc  & R@20 \\ \midrule
KV-PLM        & 35.12 & 78.91 & 36.33 & 74.71 \\
\midrule
MoMu       & 37.43 & 79.71 & 37.95 & 75.36 \\
Text2Mol      & 38.27 & 79.52 & 38.96 & 77.27 \\
MoleculeSTM    & 40.58 & 80.33 & 42.61 & 78.39 \\
MolCA & 47.71 & 82.47 & 42.71 & 80.73 \\
MolCA$^\dag$  & 49.96 & 83.61 & 44.35 & 81.71 \\
3D-MoLM  & 50.92 & 81.34 & 43.15 & 80.25 \\
3D-MoLM$^\dag$ & \underline{51.52} & \underline{83.95} & \underline{45.33} & \underline{82.18} \\
\midrule
\texttt{GeomCLIP}             & \textbf{53.50}        & \textbf{85.53}        & \textbf{48.71}        & \textbf{83.50}               \\ \bottomrule
\end{tabular}}
    \vspace{-1.7em}
 \end{table}

\section{RESULTS AND DISCUSSION}
We evaluate \texttt{GeomCLIP} on three downstream tasks: property prediction, text-molecule retrieval, and molecule captioning. Figure~\ref{fig:task} illustrates how \texttt{GeomCLIP} perform these tasks. 
\begin{table}
    \centering
    \setlength{\tabcolsep}{3.8pt}
\fontsize{9}{9}\selectfont
    \centering
    \caption{Molecule captioning results on \texttt{PubChem3D} dataset. For all methods, we utilize the decoder of MolT5-Large.}
    \vskip -1em
        \label{tab:pubchem3D_cap}
            \resizebox{0.49\textwidth}{!}{
    \begin{tabular}{lccccc} \toprule
    Model                       & BLEU-2               & BLEU-4               & ROUGE-1              & ROUGE-2              & ROUGE-L                       \\ \midrule
    MolT5                            & 26.71                 & 18.34         & 35.25                & 16.91                 & 25.37                               \\\midrule
    MoMu                               & 27.81                 & 19.61                & 35.92                & 17.33               & 26.51                         \\
    MolCA  &   28.02    & 20.68   & 35.73  & 18.06    & 27.70    \\

   3D-MoLM  &  \underline{28.71}   &  \underline{22.19}   &  \underline{36.83}   & \underline{19.86}   &  \underline{28.75}  \\
   \midrule
    \texttt{GeomCLIP}  & \textbf{31.18}    & \textbf{22.97}    & \textbf{38.05}     & \textbf{23.65}   & \textbf{32.05}     \\
    \bottomrule
    \end{tabular}}
        \vspace{-1.7em}
\end{table}


\begin{table*}[t!]
\centering
\caption{More cases of molecule captioning and correctly highlighted texts are shown in red. We observe that \texttt{GeomCLIP} is able to recognize the general class of molecule it is analyzing and identify its functional relations.}
\vskip -1em
\label{app:table-case}
\resizebox{0.8\textwidth}{!}{
    \begin{tabular}{>{\centering\arraybackslash}m{0.2\textwidth} m{0.2\textwidth} m{0.2\textwidth} m{0.2\textwidth}}
        \toprule[1pt]
        \textbf{Molecule} & \textbf{Ground Truth} & \textbf{MolT5} & \textbf{GeomCLIP}   \\
        \midrule
        \includegraphics[width=0.15\textwidth]{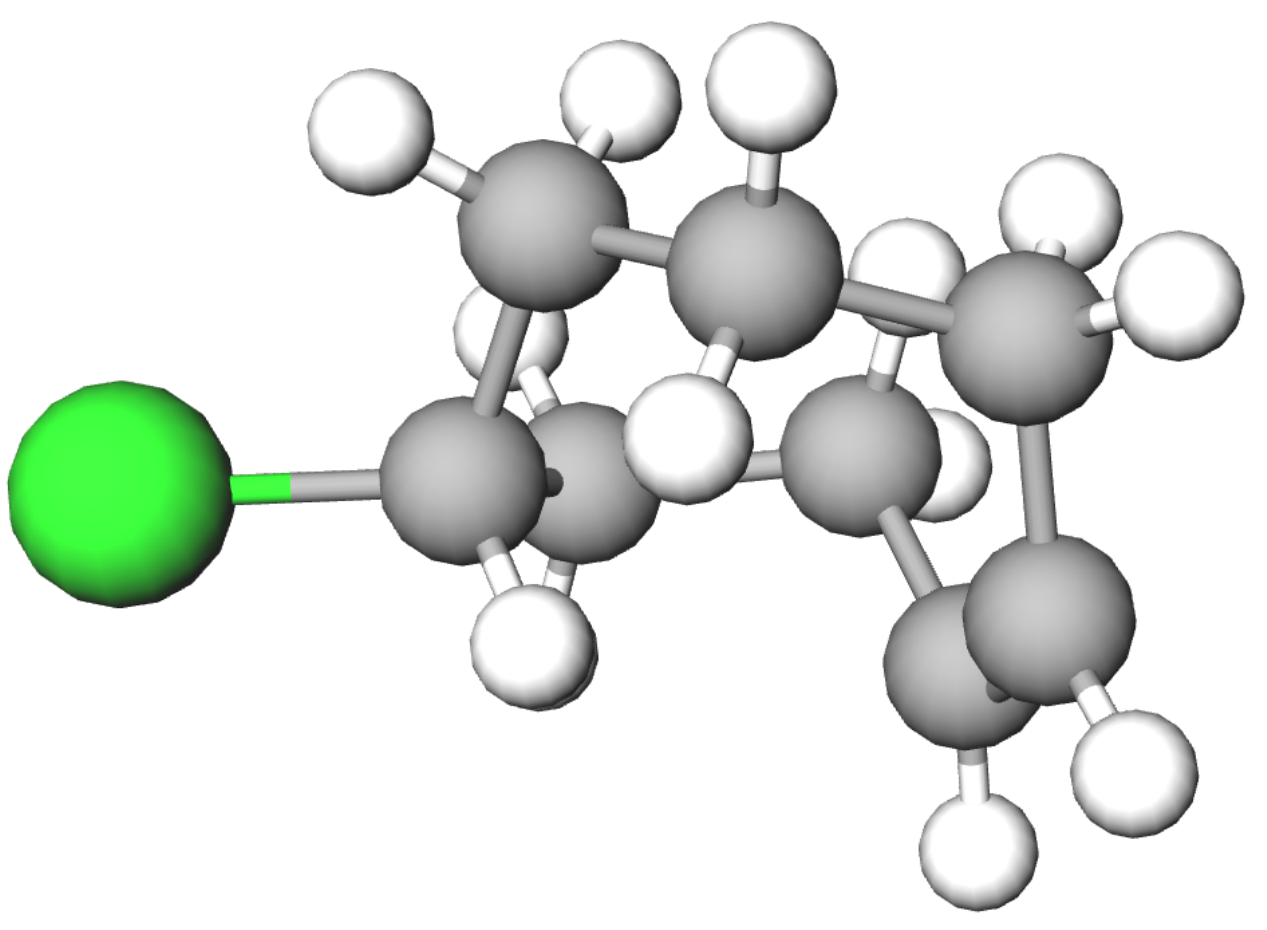} & 
        The molecule is a pyrimidone that is thymine in which the hydrogen at position 6 is substituted by a 1,3-dihydroxyisobutyl group. It is functionally related to a thymine. 
        & 
        The molecule is a \textcolor{red}{pyrimidone} that is \textcolor{red}{thymine} in which the \textcolor{red}{hydrogen} is replaced by a hydroxy group at the 5-position. It is functionally related to a uracil. 
        &
        The molecule is a \textcolor{red}{pyrimidone} that is \textcolor{red}{thymine} in which the hydrogen at position 4 is replaced by a \textcolor{red}{1,3-dihydroxyacetone group. It is functionally related to a thymine}. \\
        \midrule
        \includegraphics[width=0.18\textwidth]{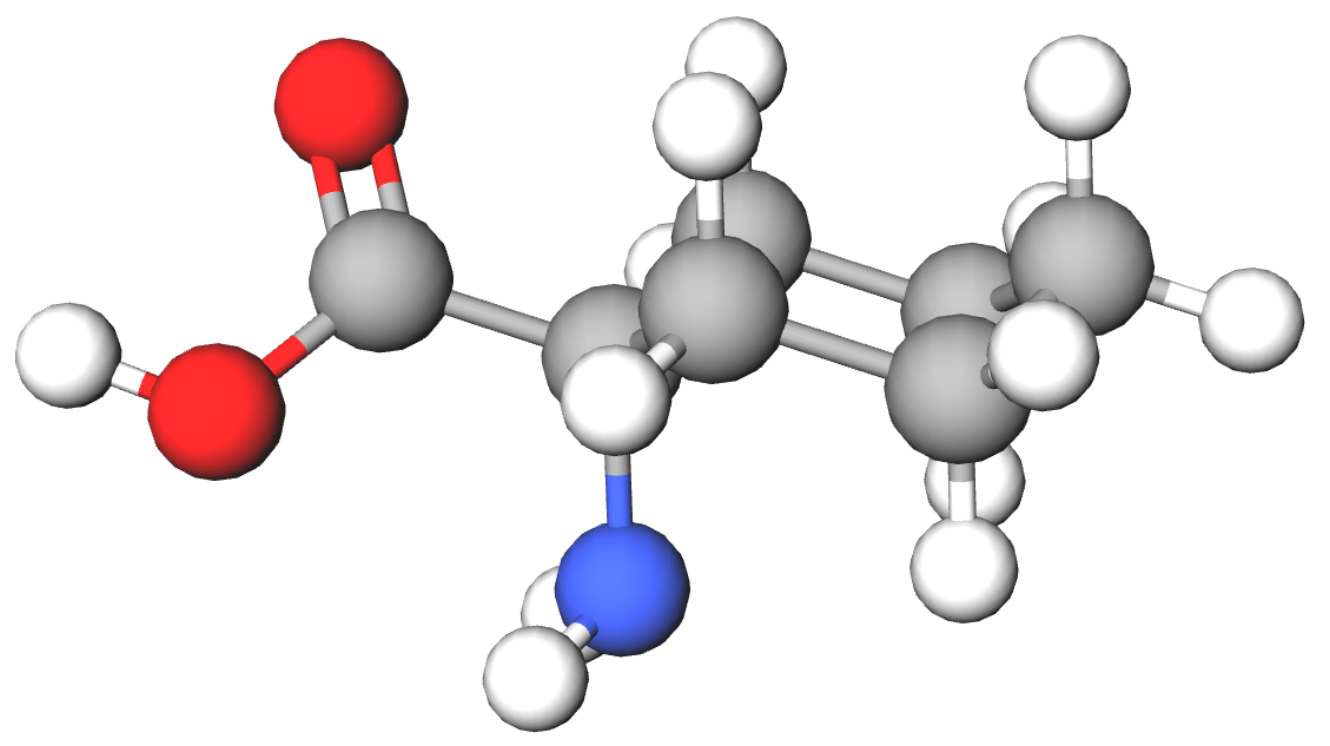} & 
        The molecule is an alpha-amino acid that is cyclohexanecarboxylic acid substituted by an amino group at position 1. It is functionally related to a cyclohexanecarboxylic acid. & The molecule is a non-proteinogenic \textcolor{red}{alpha-amino acid} that is serine in which the alcoholic hydroxy group has been formally oxidised to the corresponding formyl group. It is a non-proteinogenic alpha-amino acid and an alanine derivative. &
        The molecule is a non-proteinogenic \textcolor{red}{alpha-amino acid that is that is cyclohexanecarboxylic acid substituted by an amino group at position 1}. It is a non-proteinogenic alpha-amino acid and an alanine derivative. \textcolor{red}{It is functionally related to a cyclohexanecarboxylic acid}. \\
        \bottomrule[1pt]
    \end{tabular}}
\end{table*}

\subsection{Implementation  Details}
The pretraining experiments are conducted on four NVIDIA A100 GPUs and downstream experiments are conducted on a single NVIDIA A100 GPU. For all methods, the batch size is 64 per GPU and the gradients are accumulated for 4 steps before updating, and the representation dimension is set to 512. The learning rate is set to 0.0001 with a warm-up for the first 1,000 steps and a linear decay for the remaining steps. We use Adam~\cite{kingma2014adam} optimizer for optimization, and the weight decay is set to 0.05. 
The parameters for Uni-mol are directly borrowed from~\cite{zhou2022uni}. A small grid search is used to select the best hyperparameter for all methods.  For our \texttt{GeomCLIP}, we set temperature $\tau=0.1$ and search $\alpha$ from \{0.2, 0.4, 0.6, 0.8, 1.0\}. We select the best configuration of hyper-parameters based using the validation set.

\subsection{molecular property Prediction}
 We adopt the popular dataset: QM9~\cite{ramakrishnan2014quantum}, which is a dataset of 134K molecules consisting of 9 heavy atoms, and the 12 tasks are related to quantum properties. We follow the official  splits~\cite{wu2018moleculenet} and take 110K for training, 10K for validation, and 11K for testing. The metric is the mean absolute error (MAE).\\
\textbf{Baselines.} We consider the following baselines: 3D InfoMax~\cite{stark20223d}, GraphMVP~\cite{liu2021pre}, MoleculeSDE~\cite{liu2023group},  MoleculeJAE~\cite{chen2023molecule}, Uni-Mol~\cite{zhou2022uni} and 3D-MoLM~\cite{li2024towards}.\\
\textbf{Results.}
As shown in Table~\ref{tab:main_QM9_result}, \texttt{GeomCLIP} consistently outperforms all baselines on 12 tasks. This highlights the importance of 3D information for molecular property prediction and confirms \texttt{GeomCLIP}'s superior capability in learning 3D molecular representations. Notably, \texttt{GeomCLIP} shows significant improvements over Uni-Mol and 3D-MoLM, demonstrating the advantages of integrating ground-state 3D geometrics with their textual descriptions for property prediction.

\subsection{Zero-shot Molecule-Text Retrieval}
\vskip -0.2em
With molecule-text aligned representation space, \texttt{GeomCLIP} allows for the retrieval of molecules using texts or vice versa. We evaluate \texttt{GeomCLIP}'s performance in molecule-text retrieval on \texttt{PubChem3D}. We randomly select two subsets of 1, 500 pairs each for validation and testing. We measure retrieval performance using Accuracy and Recall@20 across the entire test set.\\
\noindent \textbf{Baseline.}
We compare \texttt{GeomCLIP} with recent baselines: MoleculeSTM~\cite{liu2023multi}, MoMu~\cite{su2022molecular}, KV-PLM~\cite{zeng2022deep}, Text2Mol~\cite{edwards2021text2mol}, MolCA~\cite{liu2023molca}, and 3D-MoLM~\cite{li2024towards}.\\
\textbf{Results.}
The results in Table~\ref{tab:retrieve_PubChem3D} reveal that \texttt{GeomCLIP} significantly outperforms existing baselines, including 1D SMILES-text models (e.g., KV-PLM) and 2D graph-text models (e.g., MoleculeSTM and MolCA). This underscores the advantage of incorporating 3D geometry in aligning the semantic spaces of molecules and texts. Furthermore, \texttt{GeomCLIP} exceeds the performance of 3D-MoLM, showcasing its ability to extract molecular features closely related to textual descriptions. The superior performance of \texttt{GeomCLIP} can be partially credited to our curated \texttt{PubChem3D} dataset, consisting of high-quality ground-state geometries. This is evidenced by 3D-MoLM$^\dag$ improved performance when retrained on \texttt{PubChem3D}, compared to it using unreal 3D geometries generated by RdKit.

\subsection{Molecule Captioning}
\vskip -0.3em
Essentially, \texttt{GeomCLIP} transforms 3D molecule presentation into underlying text space, thereby enabling it to perform a molecule-to-text generation task.  Inspired by image captioning, which integrates CLIP’s pre-trained image embeddings with GPT-2 pre-trained text generation model through a learnable mapping network. We adopt a similar strategy as shown in Figure~\ref{fig:task}, to facilitate integration with the MolT5’s~\cite{edwards2022translation} pre-trained molecule-to-text decoder, we optimize our predictor $f_{\theta}$ with next-token prediction loss. This approach lets us leverage the existing pre-trained decoder without the need for training from scratch. We use BLEU~\cite{papineni2002bleu} and ROUGE~\cite{lin2004rouge} as evaluation metrics. \\
\noindent \textbf{Baselines}. We compare with MolT5~\cite{edwards2022translation}, MoMu~\cite{su2022molecular}, Molca~\cite{liu2023molca} and 3D-MoLM~\cite{li2024towards}. \\
\textbf{Results.}
Table~\ref{tab:pubchem3D_cap} shows that our \texttt{GeomCLIP} consistently outperforms the baselines by a large margin. Specifically, it achieves improvements of up to 2.47 and 3.79 points compared to 3D-MoLM in BLEU-2 and ROUGE-2, respectively.  This showcases the effectiveness of 3D molecule-text alignment training in connecting 3D molecular representations with the input space of language models.
Table~\ref{app:table-case} shows several molecule captioning examples of different models’ outputs.  We can observe that using our \texttt{GeomCLIP} as encoder leads to more accurate descriptions of molecule structures compared to baseline MolT5.


\section{CONCLUSION}
\texttt{GeomCLIP} introduces a novel pre-training strategy that effectively combines 3D geometric information of molecules with textual descriptions, addressing the issue that corresponding texts are usually overlooked in current molecular representation learning methods. Also, by leveraging a contrastive learning approach and a denoising pre-training strategy, \texttt{GeomCLIP}-induced geometric encoder has been verified to be effective across multiple downstream applications, including property prediction, molecule-text retrieval, and molecule captioning. The creation of the new \texttt{PubChem3D} dataset which aligns geometric representations with their diverse descriptions enriches the resources that researchers can use, thus promoting future study.

\section*{Acknowledgements}
The work of Teng Xiao,  Huaisheng Zhu was supported in part by graduate research assistantships funded by grants from the National Science Foundation (2226025, 2020243) to  Vasant G. Honavar and from the National Center for Advancing Translational Sciences, and the National Institutes of Health (UL1 TR002014) to the Pennsylvania State University.

\bibliographystyle{IEEEtranS} 
\bibliography{IEEEexample}

\newpage

\end{document}